\documentclass[smallextended]{svjour3}       
\usepackage{blindtext, graphicx}
\usepackage{amsmath}
\usepackage{amssymb}
\usepackage{multirow}
\usepackage{marvosym}
\usepackage[compress]{cite}
\usepackage{subfigure}
\usepackage{color}
\usepackage[noend]{algpseudocode}
\usepackage{algorithmicx,algorithm}

\newcommand{\ie}{i.e.}
\newcommand{\etal}{\textit{et al.~}}
\begin{document}

\title{Learning a Representation with the Block-Diagonal Structure for Pattern Classification  
}


\author{He-Feng Yin$^{1,2}$         \and Xiao-Jun Wu$^{1,2*}$
\and Josef Kittler$^{3}$
         \and Zhen-Hua Feng$^{3}$
}


\institute{
\Letter\ Xiao-Jun Wu \at
              \email{xiaojun\_wu\_jnu@163.com}           
\\
\\
$^{1}$School of Internet of Things Engineering, Jiangnan University, Wuxi 214122, China
$^{2}$Jiangsu Provincial Engineering Laboratory of Pattern Recognition and Computational Intelligence, Jiangnan University, Wuxi 214122, China \and $^{3}$Centre for Vision, Speech and Signal Processing, University of Surrey, Guildford GU2 7XH, UK
}

\date{Received: date / Accepted: date}

\maketitle

\begin{abstract}
Sparse-representation-based classification (SRC) has been widely studied and developed for various practical signal classification applications. However, the performance of a SRC-based method is degraded when both the training and test data are corrupted. To counteract this problem, we propose an approach that learns Representation with Block-Diagonal Structure (RBDS) for robust image recognition. To be more specific, we first introduce a regularization term that captures the block-diagonal structure of the target representation matrix of the training data. The resulting problem is then solved by an optimizer. Last, based on the learned representation, a simple yet effective linear classifier is used for the classification task. The experimental results obtained on several benchmarking datasets demonstrate the efficacy of the proposed RBDS method. \textit{The code will be released upon the acceptance of this paper.}
\end{abstract}

\section{Introduction}
\label{intro}
In recent years, sparse representation has gained significant attention due to its successful applications in image recognition~\cite{zheng2018kernel,shao2017dynamic}, object tracking~\cite{liu2018robust}, subspace clustering~\cite{wang2016lowrank} and many other computer vision tasks~\cite{song2018dictionary, chhatrala2019sparse, song2017half}. The pioneering work in pattern recognition utilising sparse-representation-based classification (SRC) is attributed to Wright \etal\cite{wright2009robust}. SRC expresses an input test pattern as a sparse linear superposition of all the training data. Its classification is performed by checking which class conditional subset of the reconstruction coefficients produces the lowest reconstruction error. 

It has been demonstrated in~\cite{wright2009robust} that SRC is robust when the test image is occluded or corrupted, provided the training data is clean (\ie, no occlusion or corruption).  However, when the training data contains occluded or corrupted samples, the performance of SRC is degraded.  In this paper, we address the problem when both the training and test data are corrupted and present an approach that alleviates it.

To improve the robustness of SRC with respect to corrupted training data, a low-rank matrix recovery (LRMR) method has been proposed to obtain a low rank part of the corrupted image content. There are a number of previous attempts to deal with outliers. For instance, Candes \etal\cite{candes2011robust} presented the robust PCA (RPCA), which assumes that the observations lie in a single subspace such that they can be decomposed into two separate components, \ie, the low-rank normal data and a sparse noise part. However, RPCA cannot handle the situation where corrupted or outlying data are drawn from a union of multiple subspaces. To this end, Liu \etal\cite{liu2013robust} proposed a low-rank representation (LRR) method.

Based on LRMR, many approaches have been presented for robust pattern classification. Ma \etal\cite{ma2012sparse} presented a discriminative low-rank dictionary for sparse representation (DLRD\_SR). By integrating rank minimization into sparse representation for dictionary learning, this method achieved impressive face recognition results,  especially in the presence of corruption. Zhang \etal\cite{zhang2013learning}  proposed a low-rank structure representation for image classification by adding an ideal-code regularization term to the objective function. 

Recently, Li \etal\cite{li2014learninglow} advocated discriminative dictionary learning with low-rank regularization ($\textrm{D}^2$$\textrm{L}^2$$\textrm{R}^2$) for image classification that can handle training samples corrupted with large noise. $\textrm{D}^2$$\textrm{L}^2$$\textrm{R}^2$ combines the Fisher discrimination function with a low-rank constraint on the sub-dictionary to make the learned dictionary more discerning and pure. Inspired by the low rank constraint on the sub-dictionary and the ideal-code regularization term, Nguyen \etal\cite{nguyen2016discriminative} proposed a Discriminative Low-Rank Dictionary Learning (DLR\_DL) method for face recognition. 

Zheng \etal\cite{zheng2014fisher} designed a novel low rank matrix recovery algorithm with the Fisher discriminant regularization (FDLR). Wei \etal\cite{wei2014robust} introduced a constraint of structural incoherence into RPCA and presented a method called low rank matrix recovery with structural incoherence (LRSI). Based on LRSI, Yin \etal\cite{yin2016face} presented a new method that can correct the corrupted test images with a low rank projection matrix. Later Chen \etal\cite{chen2014sparse} proposed a discriminative low-rank representation (DLRR) method by incorporating structural incoherence into the framework of LRR. 

Dong \etal\cite{dong2016orthonormal} explored a discriminative orthonormal dictionary learning method for low-rank representation. Rong \etal\cite{rong2017lowrank} presented a novel low-rank double dictionary learning (LR$\textrm{D}^2$L) approach that simultaneously learns a low-rank class-specific sub-dictionary for each class and  a low-rank class-shared dictionary. Gao \etal\cite{gao2017learning} constructed a robust and discriminative low-rank representation (RDLRR) by exploiting the low rank characetristics of both the data representation and each occlusion-induced error image simultaneously. Du \etal\cite{du2018discriminative} introduced a discriminative low-rank graph preserving dictionary learning (DLRGP\_DL) method to learn a discriminative structured dictionary for sparse representation based image recognition. Recently, Wu \etal\cite{wu2018occluded} proposed a gradient direction-based hierarchical adaptive sparse and low-rank (GD-HASLR) model to solve the real-world occluded face recognition problem.

{\color{red}Though the aforementioned methods achieve encouraging results in various classification tasks, structural information content of the training data is not fully exploited.} Indeed, all recent works~\cite{li2014learning,zhang2018discriminative} indicate that utilizing structural information can achieve better performance in recognition tasks. {\color{red}In~\cite{zhang2018discriminative}, there is no dictionary learning process, \ie, the training data is directly utilized as the dictionary. This impacts on the classification performance that will be adversely affected when both the training and test data are corrupted.} 

In this paper, aiming at overcoming the above drawbacks, we propose an approach that learns a representation with block-diagonal structure (RBDS) for robust recognition. Concretely, a regularization term, which can capture the block-diagonal structure of the target representation matrix of the training data is introduced, enhancing the discriminative potential of the learned representations. Furthermore, {\color{red}we adopt an innovative strategy to solve the resulting optimization problem.} In addition, a compact dictionary is learned by our approach. 

In summary, our main contributions include:
\begin{enumerate} 
\item A new approach that learns a robust representation mirroring a block-diagonal structure is developed, which is insensitive to corruption of both training and testing images.

\item {\color{red}A compact dictionary with favourable reconstruction and discrimination properties is learned in the training stage of our proposed method.}

\item  An effective optimization technique based on the alternating direction method of multipliers (ADMM) is presented to solve the proposed problem.
\end{enumerate}

The remainder of this paper is structured as follows. Section~\ref{sec_2} reviews related work on low rank matrix recovery. In Section~\ref{sec_3},  we present our proposed approach,  with detailed optimization procedures given in Section~\ref{sec_4}. Section~\ref{sec_5} reports the experimental results on five benchmarking  datasets. Last, the conclusion is drawn in Section~\ref{sec_6}.

\section{Low Rank Matrix Recovery}
\label{sec_2}
Suppose the data matrix $\mathbf{X}$ can be decomposed into two matrices, \ie, $\mathbf{X}=\mathbf{A}+\mathbf{E}$, where $\mathbf{A}$ is a low rank matrix and $\mathbf{E}$ is the error matrix. The robust principal component analysis (RPCA) derives a low-rank matrix $\mathbf{A}$ from the corrupted data matrix $\mathbf{X}$~\cite{candes2011robust}. The objective function of RPCA is formulated as,
\begin{equation}
\label{eq:rpca}
\underset{\mathbf{A},\mathbf{E}}{\textrm{min}} \ rank(\mathbf{A})+\lambda\left \| \mathbf{E} \right \|_0,\ \textrm{s.t.} \ \mathbf{X}=\mathbf{A}+\mathbf{E}
\end{equation}
where $rank(\cdot )$ is the rank of a matrix, $\left \| \cdot  \right \|_0$ means the $\ell_0$ pseudo-norm, and $\lambda$ is a balance parameter. RPCA implicitly assumes that the underlying data structure lies in a single low rank subspace. However, this assumption is not realistic in many practical applications. Let us take face images as an example. While images of one individual tend to be drawn from the same subspace, images of distinct persons are drawn from different subspaces. Therefore, a more realistic assumption is that data samples are drawn from a union of multiple subspaces.

In order to accommodate data from multiple subspaces, Liu \etal\cite{liu2013robust} generalized the concept of RPCA and proposed a more general rank minimization problem, which is formulated as,
\begin{equation}
\label{eq:lrr}
\underset{\mathbf{Z},\mathbf{E}}{\textrm{min}} \ rank(\mathbf{Z})+\lambda\left \| \mathbf{E} \right \|_0 \ \textrm{s.t.} \ \mathbf{X}=\mathbf{D}\mathbf{Z}+\mathbf{E}
\end{equation}
where $\mathbf{D}$ is a dictionary that spans the data space.

\section{The Proposed Method}
\label{sec_3}
We first introduce the notations to be used in this paper. {\color{red}{$\mathbf{X}=[\mathbf{X}_1,\mathbf{X}_2,\cdots,\mathbf{X}_C] \in \mathbb{R}^{d \times n}$}} is the matrix of training data from $C$ classes, {\color{red}$\mathbf{X}_i \in \mathbb{R}^{d \times n_i}$} is the matrix of class $i$ with $n_i$ samples of dimension $d$ and $n=\sum_{i=1}^{C}n_i$. {\color{red}$\boldsymbol{1}_{m}=[1,\cdots,1]^T\in\mathbb{R}^{m \times 1}$ denotes the all-one vector.} Each sample in $\mathbf{X}$ can be expressed by the linear superposition of atoms in dictionary {\color{red}$\mathbf{D}$},
\begin{equation}
\label{eq:linear_rep}
\mathbf{X} = \mathbf{D}\mathbf{Z}
\end{equation}
where {\color{red}$\mathbf{D}=[\boldsymbol{d}_1,\boldsymbol{d}_2,\ldots,\boldsymbol{d}_m] \in \mathbb{R}^{d \times m}$ is the dictionary, $\boldsymbol{d}_i$ is the $i$-th atom in $\mathbf{D}$,}  and {\color{red}$\mathbf{Z}\in \mathbb{R}^{m \times n}$} is the representation matrix of the training data.
\subsection{Low-Rank and Sparse Representation}
\label{sec3_1}
It has been convincingly demonstrated in the literature that sparse representation achieves promising results in classification tasks. Similarly, it has been established that the low-rank property is a powerful  concept enabling to capture the structure information of high-dimensional data, which is robust to sparse noise.  In~\cite{zhang2013learning}, low-rank and sparse representation are combined to exploit the above two aspects. Accordingly, the problem of learning low-rank and sparse representation can be formulated as follows,
\begin{equation}
\label{eq:lrsr_ori}
\underset{\mathbf{Z},\mathbf{E}}{\textrm{min}} \ rank(\mathbf{Z})+\lambda\left \| \mathbf{E} \right \|_0+\beta\left \| \mathbf{Z} \right \|_0 \ \textrm{s.t.} \ \mathbf{X}=\mathbf{D}\mathbf{Z}+\mathbf{E}
\end{equation}
Due to the discrete properties of the rank function and the $\ell_0$-norm minimization, it is practically difficult to solve Eq.~(\ref{eq:lrsr_ori}). A common way is to replace the rank function and $\ell_0$-norm with nuclear norm and $\ell_1$-norm, respectively. Thus, Eq.~(\ref{eq:lrsr_ori}) can be reformulated as,
\begin{equation}
\label{eq:lrsr}
\underset{\mathbf{Z},\mathbf{E}}{\textrm{min}}\left \| \mathbf{Z} \right \|_*+\lambda\left \| \mathbf{E} \right \|_1+\beta\left \| \mathbf{Z} \right \|_1 \ \textrm{s.t.} \ \mathbf{X}=\mathbf{D}\mathbf{Z}+\mathbf{E}
\end{equation}
where $\lambda$ and $\beta$ control the sparsity of the noise term $\mathbf{E}$ and of the representation term $\mathbf{Z}$. The low-rank and sparse representation can be obtained by solving Eq.~(\ref{eq:lrsr}) with respect to the given dictionary $\mathbf{D}$. 

The results reported in~\cite{zhang2013learning} demonstrate the effectiveness of the low-rank and sparse representation for classification tasks. However, the solution does not focus on discriminative information. To rectify this deficiency, in the next section, we present our approach which learns a representation mirroring the Block-Diagonal Structure (RBDS) for pattern classification.
\subsection{Exploiting the Block-Diagonal Structure}
\label{sec3_2}
Several works have exploited the block-diagonal structure of the representation matrix $\mathbf{Z}$~\cite{zhang2013learning,li2014learning,zhang2018discriminative}. For example, Zhang \etal\cite{zhang2013learning} proposed a discriminative, low-rank structure framework for image classification by introducing an idealised structure as a regularization constraint.  Accordingly, the influence of the samples from the same class on an input pattern represenation is regularized to be the same. Later, Li \etal\cite{li2014learning} argued that it is unreasonable to introduce such an ideal regularization term. They presented an algorithm to learn Representation with a Classwise Block-Diagonal (RCBD) structure. Recently, Zhang \etal\cite{zhang2018discriminative} developed a discriminative block-diagonal low rank representation (BDLRR) for recognition. 

\begin{figure}[t]
\centering
\includegraphics[width = .8\textwidth]{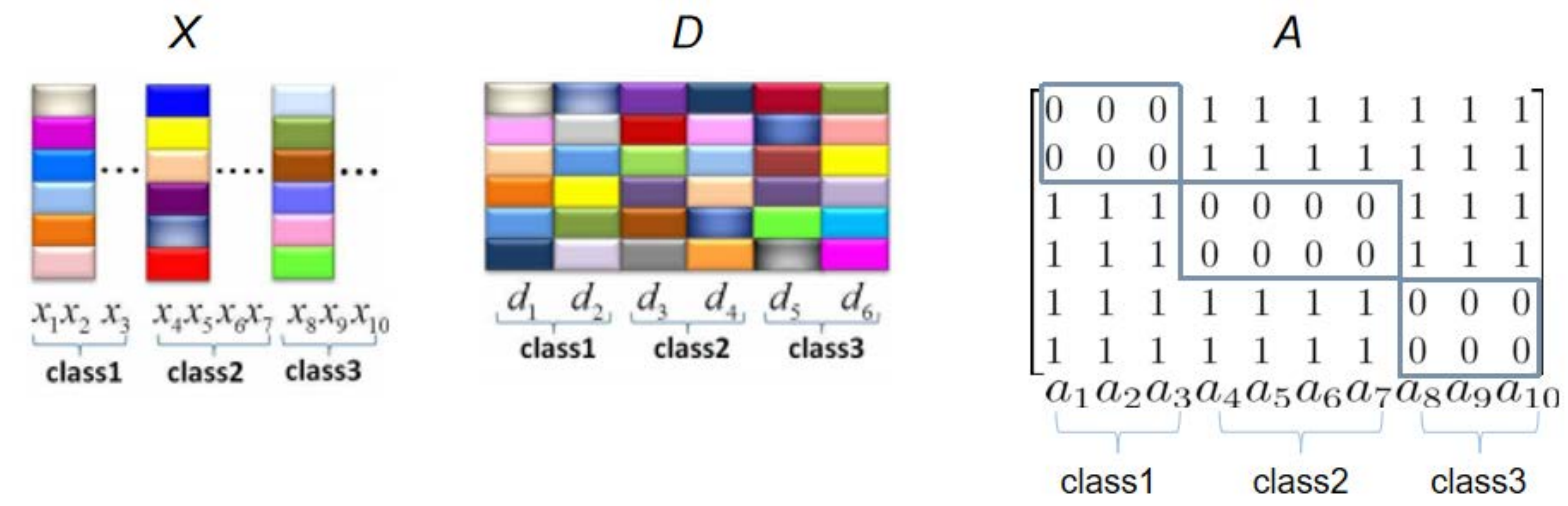}
\caption{An illustration of $\mathbf{A}$.}
\label{fig:matrix_A}
\end{figure}

Our proposed approach is similar to BDLRR. However, BDLRR does not involve dictionary learning, and this oversight inhibits the method to realise its potential when both the training and test data are corrupted. We propose an intuitive way to exploit the block-diagonal structure inherent in the training data to minimize the off-block-diagonal entries of the representation matrix. We seek to capture the block-diagonal structure by adding a regularization term {\color{red}$\left \| \mathbf{A} \odot \mathbf{Z} \right \|_F^2$}, where $\mathbf{A}$ is defined as,
\begin{equation}
\label{eq:A}
\mathbf{A}(i,j)=\left\{\begin{matrix}
0,\textrm{if} \ \boldsymbol{d}_i \ \textrm{and} \ \boldsymbol{x}_j \ \textrm{belong} \ \textrm{to} \ \textrm{the} \ \textrm{same} \ \textrm{class}
\\ 
1, \ \textrm{otherwise}
\end{matrix}\right.
\end{equation}
{\color{red}in which $\boldsymbol{d}_i$ is the $i$-th atom in dictionary $\mathbf{D}$}. An example of $\mathbf{A}$ is shown in Fig.~\ref{fig:matrix_A}.
Now, the representation learning problem with block diagonal regularization term can be formulated as follows,
\begin{equation}
\label{eq:formu_no_DL}
\underset{\mathbf{Z},\mathbf{E}}{\textrm{min}} \ \left \| \mathbf{Z} \right \|_*+\lambda\left \| \mathbf{E} \right \|_1+{\color{red}\frac{\alpha}{2}\left \|\mathbf{A} \odot \mathbf{Z} \right \|_F^2}+\beta\left \| \mathbf{Z} \right \|_1, \ \textrm{s.t.} \ \mathbf{X}=\mathbf{D}\mathbf{Z}+\mathbf{E}
\end{equation}
where $\lambda$, $\alpha$ and $\beta$ are the balancing parameters for each component, and $\left \| \cdot  \right \|_F$ denotes the Frobenius norm of a matrix.

\subsection{Dictionary Learning}
\label{sec3_3}
Compact and discriminative dictionary plays an important role in robust pattern classification, especially when both the training and test data are corrupted due to occlusion and pixel corruption. As dictionary learning has been proven to achieve promising performance~\cite{wei2014robust,ma2012sparse}, we  incorporate it into our proposed framework. Accordingly, the final formulation of our proposed approach can be stated as follows,
\begin{equation}
\label{eq:formu_DL}
\underset{\mathbf{Z},\mathbf{E},\mathbf{D}}{\textrm{min}} \ \left \| \mathbf{Z} \right \|_*+\lambda\left \| \mathbf{E} \right \|_{\color{red}1}+{\color{red}\frac{\alpha}{2}\left \| \mathbf{A} \odot \mathbf{Z} \right \|_F^2}+\beta\left \| \mathbf{Z} \right \|_1+\frac{\gamma}{2}\left \| \mathbf{D} \right \|_F^2, \ \textrm{s.t.} \ \mathbf{X}=\mathbf{D}\mathbf{Z}+\mathbf{E}
\end{equation}
where $\gamma \left \| \mathbf{D} \right \|_F^2$ is to prevent a scale change during the dictionary learning process.

\subsection{Classification Based on RBDS}
{\color{red}The outcome of  the training phase of RBDS is a dictionary, $\mathbf{D}$, and the representation matrix, $\mathbf{Z}$, of the training data $\mathbf{X}$. For the test data $\mathbf{X}_{test}$, we obtain the corresponding representation matrix $\hat{\mathbf{Z}}$ by solving Eq.~(\ref{eq:lrsr}), where $\hat{\boldsymbol{z}}_j$ is the representation vector of the $j$-th test sample.} We employ a simple linear classifier to perform recognition. The linear classifier $\mathbf{W}^*$ is designed  using the representation {\color{red}$\mathbf{Z}$} of the training data and its label matrix $\mathbf{H}$. The problem of learning $\mathbf{W}^*$ can be  formulated as follows,
\begin{equation}
\label{eq:formu_classifier}
\mathbf{W}^*=\textrm{arg }\ \underset{\mathbf{W}}{\textrm{min}}\left \| \mathbf{H}-\mathbf{W}{\color{red}\mathbf{Z}} \right \|_F^2+\eta\left \| \mathbf{W} \right \|_F^2
\end{equation}
where $\eta>0$ is a parameter. It is easy to obtain the following closed-form solution for Eq.~(\ref{eq:formu_classifier}),
\begin{equation}
\label{eq:classifier_solu}
\mathbf{W}^*=\mathbf{H}{\color{red}\mathbf{Z}}^T({\color{red}\mathbf{Z}}{\color{red}{\mathbf{Z}}}^T+\eta \mathbf{I})^{-1}
\end{equation}
Then the identity of a test sample $j$ is determined by,
\begin{equation}
\label{eq:classify_test}
i^*=\textrm{arg} \  \underset{i}{\textrm{max}} \ \mathbf{W}^*\hat{\boldsymbol{z}}_j
\end{equation}
where $i^*$ corresponds to the largest output.

\section{Optimization Algorithm}
\label{sec_4}
To solve the optimization problem Eq.~(\ref{eq:formu_DL}), we obtain the following equivalent problem by introducing two auxiliary variables $\mathbf{J}$ and $\mathbf{L}$. Then Eq.~(\ref{eq:formu_DL}) can be rewritten as,
\begin{equation}
\label{eq:equ_formu}
\begin{split}
&\underset{\mathbf{Z},\mathbf{J},\mathbf{L},\mathbf{E},\mathbf{D}}{\textrm{min}} \ \left \| \mathbf{J} \right \|_*+\lambda\left \| \mathbf{E} \right \|_{\color{red}1}+{\color{red}\frac{\alpha}{2}\left \| \mathbf{A}\odot \mathbf{Z} \right \|_F^2}+\beta\left \| \mathbf{L} \right \|_1+\frac{\gamma}{2}\left \| \mathbf{D} \right \|_F^2,  \\
&\textrm{s.t.} \ \mathbf{X}=\mathbf{D}\mathbf{Z}+\mathbf{E},\mathbf{Z}=\mathbf{J},\mathbf{Z}=\mathbf{L}
\end{split}
\end{equation}
which can be solved based on the Augmented Lagrange Multiplier (ALM) method~\cite{{lin2011linearized}}. The augmented Lagrangian function of Eq.~(\ref{eq:equ_formu}) is defined as follows,
\begin{equation}
\label{eq:augment}
\begin{split}
 &\Lambda (\mathbf{Z},\mathbf{J},\mathbf{L},\mathbf{E},\mathbf{D},\mathbf{Y}_1,\mathbf{Y}_2,\mathbf{Y}_3,\mu)= \left \| \mathbf{J} \right \|_*+\lambda\left \| \mathbf{E} \right \|_{\color{red}1}+{\color{red}\frac{\alpha}{2}\left \|\mathbf{A}\odot \mathbf{Z} \right \|_F^2}+\beta\left \| \mathbf{L} \right \|_1 \\ &+\frac{\gamma}{2}\left \| \mathbf{D} \right \|_F^2+
<\mathbf{Y}_1,\mathbf{X}-\mathbf{D}\mathbf{Z}-\mathbf{E}>+<\mathbf{Y}_2,\mathbf{Z}-\mathbf{J}>+<\mathbf{Y}_3,\mathbf{Z}-\mathbf{L}> \\ &+\frac{\mu}{2}(\left \| \mathbf{X}-\mathbf{D}\mathbf{Z}-\mathbf{E} \right \|_F^2+\left \| \mathbf{Z}-\mathbf{J} \right \|_F^2+\left \| \mathbf{Z}-\mathbf{L} \right \|_F^2) 
\end{split}
\end{equation}
where $<\mathbf{A},\mathbf{B}>=trace(\mathbf{A}^t\mathbf{\mathrm{}B})$, $\mathbf{Y}_1$, $\mathbf{Y}_2$ and $\mathbf{Y}_3$ are Lagrange multipliers and $\mu > 0$ is a penalty parameter. The optimization of Eq.~(\ref{eq:augment}) can be solved iteratively by updating $\mathbf{J}$, $\mathbf{Z}$, $\mathbf{L}$, $\mathbf{E}$ and $\mathbf{D}$ once at a time. The detailed updating procedures are presented as follows.

$\textit{Updating}$ $\mathbf{J}$: Fix the other variables and update $\mathbf{J}$ by solving the following problem,
\begin{equation}
\label{eq:solve_J}
\begin{split}
\mathbf{J}^{k+1}&=\textrm{arg }\ \underset{\mathbf{J}}{\textrm{min}} \ \left \| \mathbf{J} \right \|_*+<\mathbf{Y}_2^k,\mathbf{Z}^k-\mathbf{J}>+\frac{\mu^k}{2}\left \| \mathbf{Z}^k-\mathbf{J} \right \|_F^2\\
&=\textrm{arg }\ \underset{\mathbf{J}}{\textrm{min}} \ \frac{1}{\mu^k}\left \| \mathbf{J} \right \|_*+\frac{1}{2}\left \| \mathbf{J}-(\mathbf{Z}^k+\frac{\mathbf{Y}_2^k}{\mu^k}) \right \|_F^2\\
&=\mathbf{U}S_{\frac{1}{\mu^k}}[\mathbf{\Sigma} ]\mathbf{V}^T
    \end{split}
\end{equation}
where $(\mathbf{U},\mathbf{\Sigma},\mathbf{V}^T)=SVD(\mathbf{Z}^k+\mathbf{Y}_2^k/\mu^k) $ and $S_{\varepsilon }[\cdot]$ is the soft-thresholding (shrinkage) operator defined as follows~\cite{lin2011linearized},
\begin{equation}
\label{eq:soft-thres}
S_{\varepsilon }[x]=\left\{\begin{matrix}
x-\varepsilon, \textrm{if} \ x>\varepsilon\\ 
x+\varepsilon, \textrm{if} \ x<-\varepsilon\\ 
0,\textrm{otherwise}
\end{matrix}\right. 
\end{equation}
$\textit{Updating}$ $\mathbf{Z}$: To update $\mathbf{Z}$, we fix all the variables other than $\mathbf{Z}$ and solve the following problem:
\begin{equation}
\label{eq:solve_Z_ori}
\begin{split}
&\mathbf{Z}^{k+1}=\textrm{arg}\ \underset{\mathbf{Z}}{\textrm{min}} \ {\color{red}\frac{\alpha}{2}\left \| \mathbf{A}\odot \mathbf{Z} \right \|_F^2}+\frac{\mu^k}{2}\left \| \mathbf{X}-\mathbf{D}^k\mathbf{Z}-\mathbf{E}^k+\frac{\mathbf{Y}_1^k}{\mu^k} \right \|_F^2+\\ &\frac{\mu^k}{2}\left \| \mathbf{Z}-\mathbf{J}^{k+1}+\frac{\mathbf{Y}_2^k}{\mu^k} \right \|_F^2 +\frac{\mu^k}{2}\left \| \mathbf{Z}-\mathbf{L}^k+\frac{\mathbf{Y}_3^k}{\mu^k} \right \|_F^2
\end{split}
\end{equation}
which is equivalent to
\begin{equation}
\label{eq:solve_Z}
\begin{split}
&\mathbf{Z}^{k+1}=\textrm{arg}\ \underset{\mathbf{Z}}{\textrm{min}} \ \frac{\alpha}{2}\left \| \mathbf{Z}-\mathbf{R} \right \|_F^2+\frac{\mu^k}{2}\left \| \mathbf{X}-\mathbf{D}^k\mathbf{Z}-\mathbf{E}^k+\frac{\mathbf{Y}_1^k}{\mu^k} \right \|_F^2\\&+\frac{\mu^k}{2}\left \| \mathbf{Z}-\mathbf{J}^{k+1}+\frac{\mathbf{Y}_2^k}{\mu^k} \right \|_F^2 +\frac{\mu^k}{2}\left \| \mathbf{Z}-\mathbf{L}^k+\frac{\mathbf{Y}_3^k}{\mu^k} \right \|_F^2
\end{split}
\end{equation}
where $\mathbf{R}={\color{red}\mathbf{M}}\odot \mathbf{Z}^k$ and $\mathbf{M}=\boldsymbol{1}_m\boldsymbol{1}_n^T-\mathbf{A}$. Eq.~(\ref{eq:solve_Z}) has a closed-form solution, given by
\begin{equation}
\label{eq:Z_closed}
\mathbf{Z}^{k+1}=[(\mathbf{D}^k)^T\mathbf{D}^k+(\frac{\alpha}{\mu^k}+2)\mathbf{I}]^{-1}[(\mathbf{D}^k)^T(\mathbf{X}-\mathbf{E}^k)+\mathbf{J}^{k+1}+\mathbf{L}^k+\frac{\alpha \mathbf{R}+(\mathbf{D}^k)^T\mathbf{Y}_1^k-\mathbf{Y}_2^k-\mathbf{Y}_3^k}{\mu^k}]  
\end{equation}
$\textit{Updating}$ $\mathbf{L}$: When we fix the other variables, Eq.~(\ref{eq:augment})  degenerates into a function of $\mathbf{L}$, that is,
\begin{equation}
\label{eq:solve_L}
\begin{aligned} 
\mathbf{L}^{k+1} & =\textrm{arg} \ \underset{\mathbf{L}}{\textrm{min}} \ \frac{\beta}{\mu^k}\left \| \mathbf{L} \right \|_1+\frac{1}{2}\left \| \mathbf{L}-(\mathbf{Z}^{k+1}+\frac{\mathbf{Y}_3^k}{\mu^k}) \right \|_F^2 \\
   & =S_{\frac{\beta}{\mu^k}}[\mathbf{Z}^{k+1}+\frac{\mathbf{Y}_3^k}{\mu^k}]
\end{aligned}
\end{equation}
$\textit{Updating}$ $\mathbf{E}$: To update $\mathbf{E}$, we minimize Eq.~(\ref{eq:augment}) and fix all the variables other than $\mathbf{E}$, which leads to
\begin{equation}
\label{eq:solve_E}
\begin{aligned} 
\mathbf{E}^{k+1} & =\textrm{arg }\ \underset{\mathbf{E}}{\textrm{min}} \ \frac{\lambda}{\mu^k}\left \| \mathbf{E} \right \|_1+\frac{1}{2}\left \| \mathbf{E}-(\mathbf{X}-\mathbf{D}^k\mathbf{Z}^{k+1}+\frac{\mathbf{Y}_1^k}{\mu^k}) \right \|_F^2 \\
   & =S_{\frac{\lambda}{\mu^k}}[\mathbf{X}-\mathbf{D}^k\mathbf{Z}^{k+1}+\frac{\mathbf{Y}_1^k}{\mu^k}]
\end{aligned}
\end{equation}
$\textit{Updating}$ $\mathbf{D}$: When the other variables are fixed, optimising Eq.~(\ref{eq:augment}) with respect to $\mathbf{D}$ boils down to the following problem,
\begin{equation}
\label{eq:solve_D}
\begin{split}
&\mathbf{D}^{k+1}=\textrm{arg} \ \underset{\mathbf{D}}{\textrm{min}} \ \frac{\gamma}{2}\left \| \mathbf{D} \right \|_F^2+<\mathbf{Y}_1^k,\mathbf{X}-\mathbf{D}\mathbf{Z}^{k+1}-\mathbf{E}^{k+1}>\\
&+\frac{\mu^k}{2}\left \| \mathbf{X}-\mathbf{D}\mathbf{Z}^{k+1}-\mathbf{E}^{k+1} \right \|_F^2    
\end{split}
\end{equation}
which has a closed-form solution as follows,
\begin{equation}
\label{eq:D_solution}
\mathbf{D}^{k+1}=[\frac{\mathbf{Y}_1^k(\mathbf{Z}^{k+1})^T}{\mu^k}-(\mathbf{E}^{k+1}-\mathbf{X})(\mathbf{Z}^{k+1})^T](\frac{\gamma}{\mu^k}\mathbf{I}+\mathbf{Z}^{k+1}(\mathbf{Z}^{k+1})^T)^{-1}
\end{equation}
The detailed procedures for solving Eq.~(\ref{eq:augment}) are presented in Algorithm 1.
\begin{algorithm}[htp]
\caption{$:$ Solving Eq.~(\ref{eq:augment}) by Inexact ALM}
\label{alg:algorithm1}
\begin{algorithmic}
\Require
{\color{red}Training} data matrix $\mathbf{X}$; Paramters $\lambda$, $\alpha$, $\beta$ and $\gamma$
\Ensure
$\mathbf{Z}$, $\mathbf{D}$ and $\mathbf{E}$ \\
Initialize: $\mathbf{Z}^0$=0, $\mathbf{J}^0$=0, $\mathbf{L}^0$=0, $\mathbf{E}^0$=0, $\mathbf{Y}_1^0$=0, $\mathbf{Y}_2^0$=0, $\mathbf{Y}_3^0$=0, $\mu^0$=$10^{-5}$, $\mu_{max}$=$10^{8}$, $\rho$=1.1, $\varepsilon$=$10^{-6}$
\While{not converged} 
\State Update $\mathbf{J}$ using $(\ref{eq:solve_J})$
\State Update $\mathbf{Z}$ using $(\ref{eq:Z_closed})$
\State Update $\mathbf{L}$ using $(\ref{eq:solve_L})$
\State Update $\mathbf{E}$ using $(\ref{eq:solve_E})$
\State Update $\mathbf{D}$ using $(\ref{eq:D_solution})$
\State Update the multipliers:
\State $\mathbf{Y}_1^{k+1}=\mathbf{Y}_1^{k}+\mu^k(\mathbf{X}-\mathbf{D}^{k+1}\mathbf{Z}^{k+1}-\mathbf{E}^{k+1})$
\State $\mathbf{Y}_2^{k+1}=\mathbf{Y}_2^{k}+\mu^k(\mathbf{Z}^{k+1}-\mathbf{J}^{k+1})$
\State $\mathbf{Y}_3^{k+1}=\mathbf{Y}_3^{k}+\mu^k(\mathbf{Z}^{k+1}-\mathbf{L}^{k+1})$
\State Update $\mu$
\State $\mu^{k+1}=\textrm{min}(\mu_{max},\rho\mu^k)$
\State Check the convergence conditions:
\State $\left \| \mathbf{Z}^{k+1}-\mathbf{J}^{k+1} \right \|_{\infty}<\varepsilon$, $\left \| \mathbf{Z}^{k+1}-\mathbf{L}^{k+1} \right \|_{\infty}<\varepsilon$ and $\left \| \mathbf{X}-\mathbf{D}^{k+1}\mathbf{Z}^{k+1}-\mathbf{E}^{k+1} \right \|_{\infty}<\varepsilon$
\EndWhile
\end{algorithmic}
\end{algorithm}

\section{Experimental Results}
\label{sec_5}
\begin{figure}
\centering
\subfigure[Example images from the Extended Yale B database]{
\begin{minipage}[b]{0.7\textwidth}
\includegraphics[width=\textwidth]{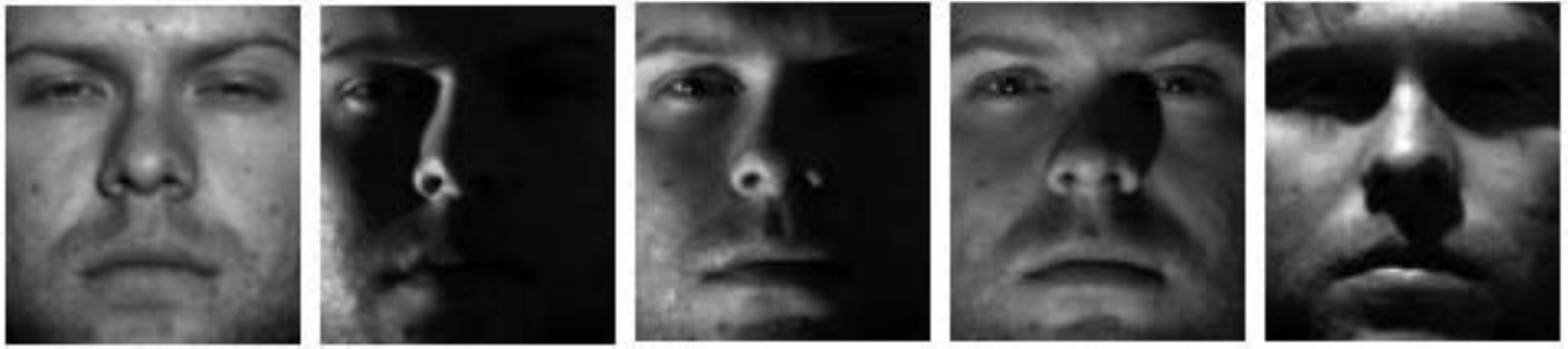}
\label{fig:EYaleB}
\end{minipage}
}

\subfigure[Example images from the AR database]{
\begin{minipage}[b]{0.7\textwidth}
\includegraphics[width=\textwidth]{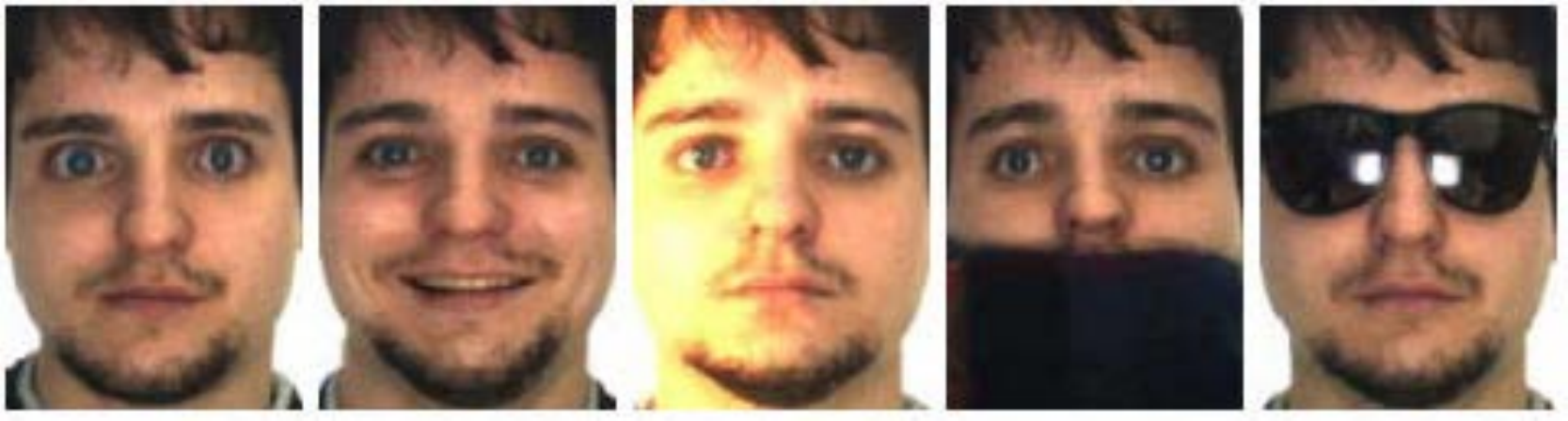}
\label{fig:AR}
\end{minipage}
}

\subfigure[Example images from the ORL database]{
\begin{minipage}[b]{0.7\textwidth}
\includegraphics[width=\textwidth]{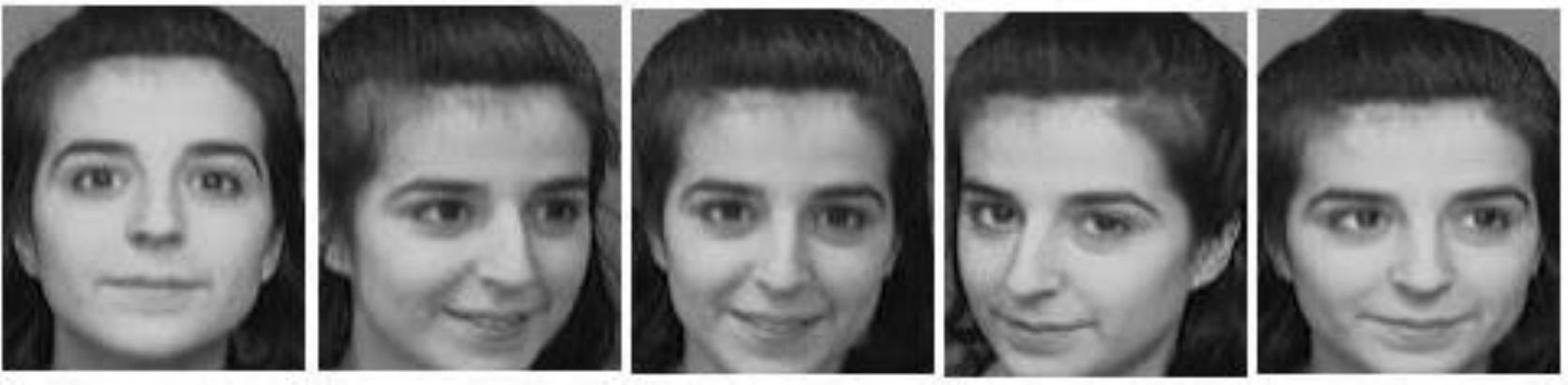}
\label{fig:ORL}
\end{minipage}
}

\subfigure[Example images from the LFW database]{
\begin{minipage}[b]{0.7\textwidth}
\includegraphics[width=\textwidth]{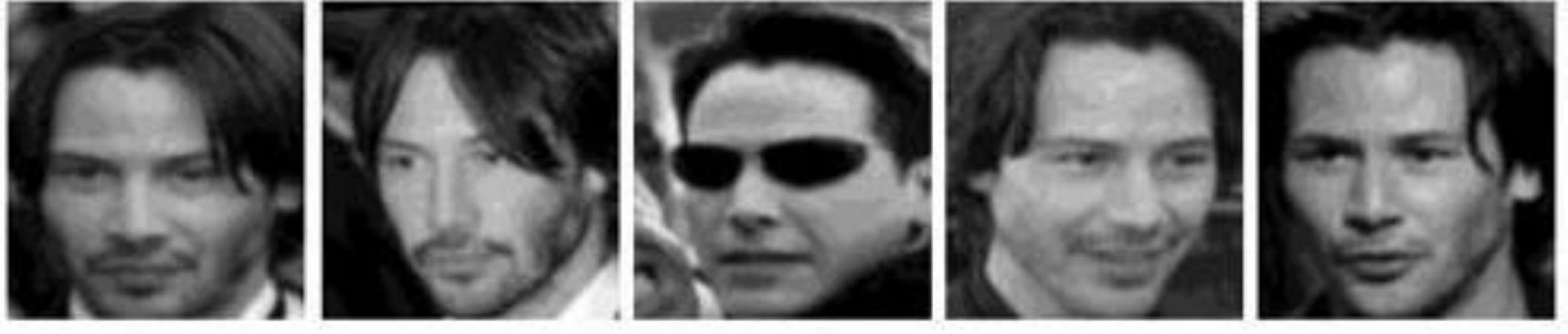}
\label{fig:LFW}
\end{minipage}
}

\subfigure[Example images from the Scene 15 dataset]{
\begin{minipage}[b]{0.7\textwidth}
\includegraphics[width=\textwidth]{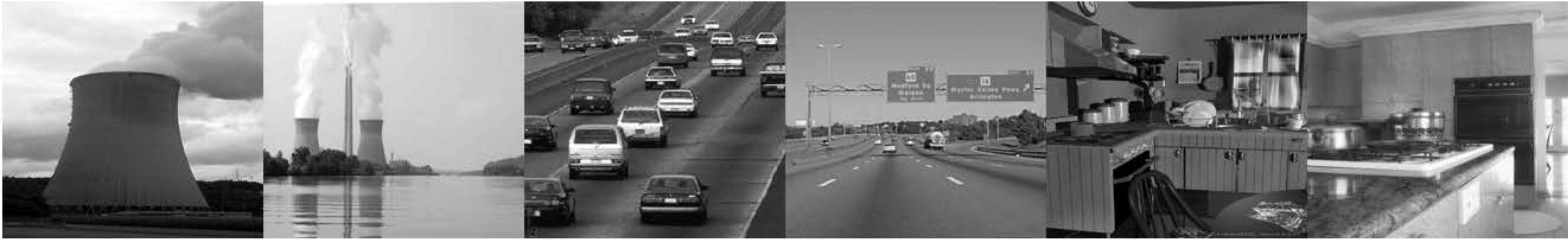}
\label{fig:Scene15}
\end{minipage}
}
\caption{Examples of the benchmarking datasets: (a) the Extended Yale B database with illumination variations; (b) the AR database with appearance variations in illumination, expression and occlusion; (c) the ORL database with expression and pose variations; (d) the LFW database including variations in pose, illumination and occlusion; (e) the Scene 15 dataset consisting of various images for a specific scene.}
\label{fig:example_imgs}
\end{figure}

The proposed RBDS is evaluated on five publicly available databases: Extended Yale B~\cite{georghiades2001from}, AR~\cite{martinez1998ar}, ORL~\cite{samaria1994orl}, LFW~\cite{huang2007labeled} and Scene 15 dataset~\cite{lazebnik2006beyond}. Example images from these databases are shown in Fig.~\ref{fig:example_imgs}. For the first four face databases, we deal with training and test images being corrupted by factors such as illumination variation, expression changes, pose variation, occlusion and uniformly distributed noise. Similar factors affect the last dataset whcih is concerned with scene classification. The proposed approach is compared with related low rank and dictionary learning methods, including {\color{red}BDLRR~\cite{zhang2018discriminative}, $\textrm{D}^2$$\textrm{L}^2$$\textrm{R}^2$~\cite{li2014learninglow}}, SLRR~\cite{zhang2013learning}, DLRD\_SR~\cite{ma2012sparse}, LRSI~\cite{wei2014robust}, RPCA~\cite{candes2011robust}, FDDL~\cite{yang2011fisher}, LLC~\cite{wang2010locality}, CRC~\cite{zhang2011sparse}, SR~\cite{wright2009robust} and SVM. It should be noted that $\textbf{SRW}$ indicates the case when all the training data is used as the dictionary. $\textbf{SRS}$ has the number of atoms as our proposed RBDS. In order to comprehensively evaluate the role of dictionary learning and the effect of the block diagonal structure term, our approach (RBDS) is compared with its two special cases: LRRS\_BD and LRRS, the objective functions of which are shown in Eqs.~(\ref{eq:formu_no_DL}) and~(\ref{eq:lrsr}), respectively. Each experiment is repeated ten times and the average recognition results are reported for all approaches. 

\subsection{Extended Yale B Database}
\label{sec:5-1}
The Extended Yale B database has 2414 frontal-face images of 38 subjects. The images of size 192$\times$168 were taken under  laboratory-controlled lighting conditions. There are between 59 and 64 images for each person. Following the experimental protocol in RCBD~\cite{li2014learning}, we evaluate our approach on images down sampled by factors 1/2, 1/4, 1/8, and the resulting feature dimensions are 8064, 2016 and 504, respectively. We randomly select $N_c$ images of each person as the training set ($N_c$ = 8, 32), and the remaining ones as the test set. When there are 8 training images of each subject, the learned dictionary has 5 atoms for each class. When $N_c$=32, the learned dictionary has 20 atoms per class. 

The results of different methods obtained on the Extended Yale B database are reported in Table~\ref{table:EYaleB}. According to the table, our RBDS method achieves the best performance in most cases, even when only a small number of training samples are available. Furthermore, the experimental results indicate that RBDS can handle the challenges of illumination and expression changes. The proposed method outperforms SLRR by {\color{red}0.7}\% with 8 training images per person, and by {\color{red}5.2}\%  with 32 training images per person on average. Our approach achieves a significant performance gain in the case of 32 training images per person.
\begin{table}[t]
\centering
\caption{Recognition accuracy in (\%) on the Extended Yale B database}
\label{table:EYaleB}
\begin{tabular}{ccccccc}
\hline
No. per class & \multicolumn{3}{c}{$N_c$=8}                      & \multicolumn{3}{c}{$N_c$=32}                     \\ \hline
Sample Rate   & 1/8           & 1/4           & 1/2           & 1/8           & 1/4           & 1/2           \\ \hline
\textbf{RBDS} & {\color{red}\textbf{80.3}} & {\color{red}82.8} & {\color{red}83.3} & {\color{red}\textbf{97.2}} & {\color{red}\textbf{98.7}} & {\color{red}\textbf{98.9}} \\
LRRS\_BD       & 76.0          & 79.7          & 81.1          & 96.7          & 97.9          & 98.0          \\
LRRS          & 75.3          & 78.6          & 79.5          & 96.8          & 96.9          & 97.7          \\
{\color{red}BDLRR~\cite{zhang2018discriminative}} & {\color{red}73.4}  & {\color{red}75.1} & {\color{red}77.6 }& {\color{red}95.8 }& {\color{red}96.8} & {\color{red}97.5 }\\
{\color{red}$\textrm{D}^2$$\textrm{L}^2$$\textrm{R}^2$~\cite{li2014learninglow}}  & {\color{red}79.3} & {\color{red}82.8} & {\color{red}\textbf{84.0}} & {\color{red} 96.1}& {\color{red}97.2} & {\color{red}97.5} \\
SLRR~\cite{zhang2013learning}          & 76.6          & \textbf{83.7}          & 83.8          & 89.9          & 93.6          & 95.7          \\
LRSI~\cite{wei2014robust}          & 73.3          & 80.9          & 80.8          & 89.5          & 93.5          & 94.5          \\
RPCA~\cite{candes2011robust}          & 74.6          & 78.3          & 80.2          & 85.6          & 90.7          & 94.1          \\
SRW~\cite{wright2009robust}           & 79.3          & 83.0          & 83.8          & 87.2          & 89.5          & 90.7          \\
SRS~\cite{wright2009robust}           & 75.3          & 78.9          & 80.1          & 84.4          & 85.7          & 85.9          \\ 
LLC~\cite{wang2010locality}           & 65.7          & 70.6          & 76.1          & 76.4          & 80.0          & 85.6          \\ \hline
\end{tabular}
\end{table}

\subsection{Evaluation on Occluded Faces}
\label{sec:5-2}
The AR database consists of over 4000 images of 126 subjects. For each individual, 26 images are taken in different conditions in two separate sessions. There are 13 images from each session, including 3 images with sunglasses, another 3 with scarves and the remaining 7 show different illumination and expression changes. The resolution of each image is 165$\times$120.

In our experiments, we use a subset of the AR database, which contains 50 male and 50 female subjects. Following the experimental protocol in RCBD~\cite{li2014learning}, we convert the color images into gray scale and down-sample them by a factor of 1/3, resulting in the dimensionality of 2200. We consider the following three scenarios:

\textit{1) Sunglasses}: We first investigate the effect of occlued samples by sunglasses, which affect about 20\% of the face image. We use seven neutral images plus one image with sunglasses (randomly chosen) from session 1 for training (eight training images per class), and the remaining neutral images (all from Session 2) and the rest of the images with sunglasses (two taken from Session 1 and three from Session 2) for testing (twelve test images per class).

\textit{2) Scarf}: Here we replace the images with sunglasses in the above scenario by images with a scarf.

\textit{3) Mixed (Sunglasses + Scarf)}: In the last scenario, the training samples may be occluded either by sunglasses or scarves, which is more challenging than the above two scenarios. Seven neutral images, two corrupted images (one with sunglasses and one with scarf) from session 1 are used for training (nine training images per class), and the remaining ones are used for testing (seventeen test images per class).

Similar to RCBD~\cite{li2014learning}, a compact dictionary with 5 atoms per class is learned under different scenarios. Table~\ref{table:AR} summarizes the experimental results on the AR database. Our approach consistently performs best and its accuracy gains over {\color{red}BDLRR} are {\color{red}4.9}\% for the sunglasses scenario, {\color{red}4.9}\% for the scarf scenario, and {\color{red}5.9}\% for the mixed scenario, respectively. As expected, LRRS\_BD is inferior to RBDS, which demonstrates that a high quality dictionary is needed for learning a discriminative representation when both training and test images are corrupted.
\begin{table}[t]
\centering
\caption{Recognition accuracy (\%) on the AR database}
\label{table:AR}
\begin{tabular}{cccc}
\hline
Scenario      & Sunglasses    & Scarf         & Mixed         \\ \hline
\textbf{RBDS} & {\color{red}\textbf{95.5}} & {\color{red}\textbf{93.3}} & {\color{red}\textbf{93.7}} \\
LRRS\_BD       & 90.6          & 86.8          & 85.7          \\
LRRS          & 89.2          & 85.2          & 85.6          \\
{\color{red}BDLRR~\cite{zhang2018discriminative}} & {\color{red}90.6}  & {\color{red}88.4} & {\color{red}87.8 }\\
{\color{red}$\textrm{D}^2$$\textrm{L}^2$$\textrm{R}^2$~\cite{li2014learninglow}}  & {\color{red}89.3} & {\color{red}84.1} & {\color{red}82.7} \\
SLRR~\cite{zhang2013learning}          & 87.3          & 83.4          & 82.4          \\
LRSI~\cite{wei2014robust}          & 84.9          & 76.4          & 80.3          \\
RPCA~\cite{candes2011robust}          & 83.2          & 75.8          & 78.9          \\
SRW~\cite{wright2009robust}           & 86.8          & 83.2          & 79.2          \\
SRS~\cite{wright2009robust}           & 82.1          & 72.6          & 65.5          \\ 
LLC~\cite{wang2010locality}           & 65.3          & 59.2          & 59.9          \\ \hline
\end{tabular}
\end{table}

\subsection{Evaluation on Pixel Corruption}
\label{sec:5-3}
\begin{figure}[t]
\centering
\includegraphics[trim={0mm 0mm 0mm 0mm},clip, width = .8\textwidth]{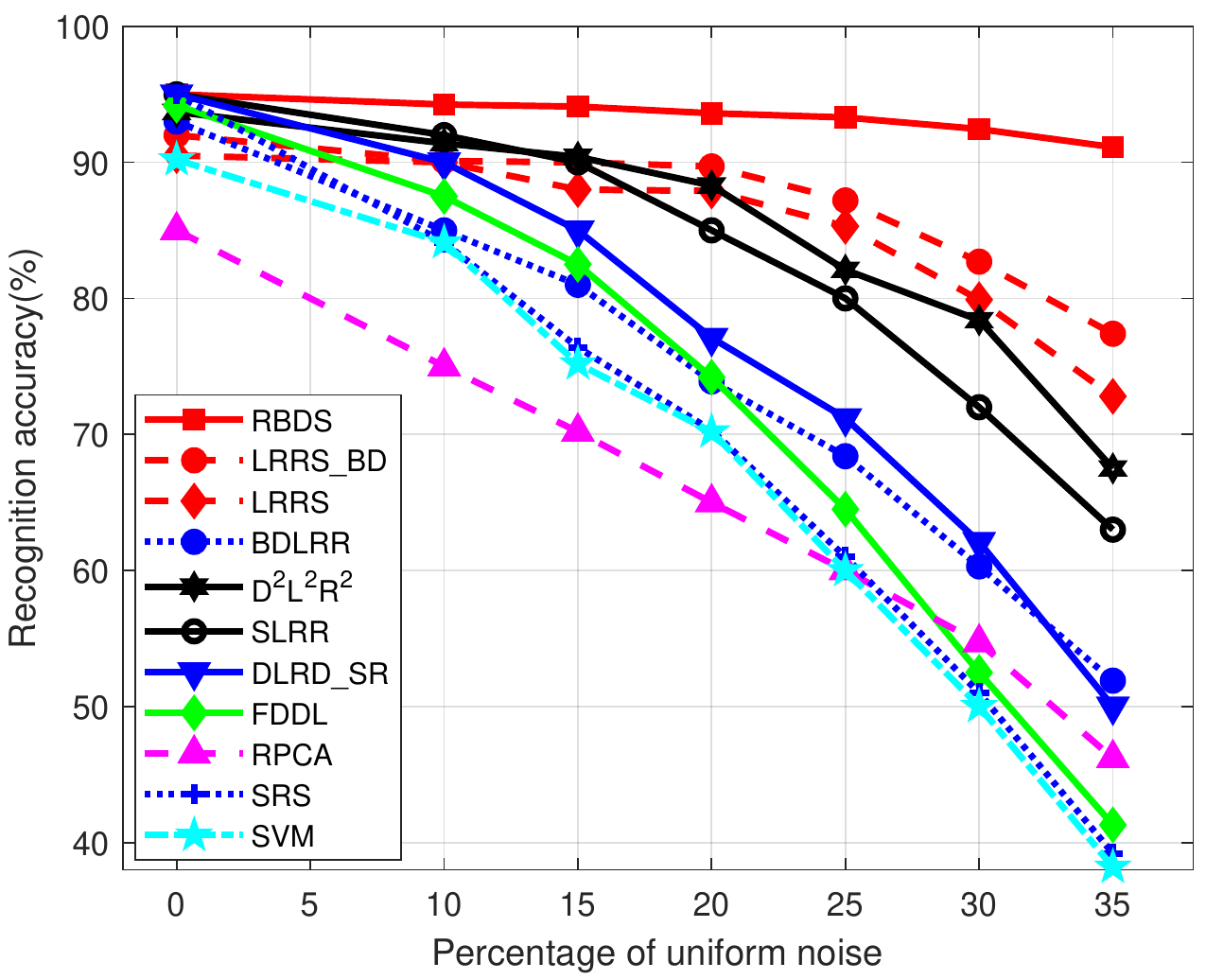}
\caption{Recognition accuracy on the AR database with different levels of pixel noise.}
\label{fig:AR_pixel_corr}
\end{figure}

In this section, we evaluate the proposed method on the AR database with different levels of corruption. First, we select 7 neutral images with illumination and expression changes from Session 1 for training, and the other 7 neutral images from Session 2 for testing. A certain percentage of randomly selected pixels from both training and test images are replaced by noise uniformly distributed between the minimal and maximal pixel value. The number of dictionary atoms per class is set to 7. The recognition accuracy is plotted under different levels of corruption in Fig.~\ref{fig:AR_pixel_corr}. Our approach outperforms {\color{red}$\textrm{D}^2$$\textrm{L}^2$$\textrm{R}^2$} by {\color{red}8.8}\% on average. Fig.~\ref{fig:AR_pixel_corr} demonstrates that the proposed RBDS consistently outperforms all the other approaches for all levels of pixel corruption.

\subsection{Evaluation on Block Occlusion}
\label{sec:5-4}
To further verify the performance of different methods in tackling random block occlusion, an experiment is carried out on the ORL database. This database has 400 images of 40 individuals. The images were taken at different times, with lighting variation, facial expression and pose changes. We crop and normalize each image to 28$\times$23 pixels. For each subject, a half of the images are randomly selected as training samples, and the remaining ones serve as test samples. We replace a randomly located block of each image with an unrelated random image. The experiments are conducted for different degrees of block occlusion.

Table~\ref{tabel:ORL} presents the recognition accuracy for different levels of occlusion on the ORL database. Our approach (RBDS) achieves the best performance and outperforms DLRD\_SR by {\color{red}1.9}\% on average. Our approach shows high robustness to the severe corruption posed by block occlusion, and achieves {\color{red}3.4}\% improvement in the case of 40\% block noise. Moreover, thanks to the block-diagonal structure term, LRRS\_BD achieves better results than LRRS.  However, LRRS\_BD  is inferior to our proposed RBDS since it  does not involve the dictionary learning process.
\begin{table}[]
\centering
\caption{Recognition accuracy in (\%) on the ORL database with block occlusion}
\label{tabel:ORL}
\begin{tabular}{lllllll}
\hline
Noise percent & 0             & 10            & 20            & 30            & 40            & 50            \\ \hline
\textbf{RBDS} & {\color{red}96.0}          & {\color{red}\textbf{94.7}} & {\color{red}\textbf{91.9}} & {\color{red}\textbf{89.2}} & {\color{red}\textbf{80.1}} & {\color{red}\textbf{73.5}} \\
LRRS\_BD      & 96.1          & 94.7          & 91.1          & 86.0          & 79.9          & 70.6          \\
LRRS          & 96.1          & 93.6          & 89.8          & 84.8          & 77.7          & 70.9          \\
{\color{red}BDLRR~\cite{zhang2018discriminative}} & {\color{red}95.3}  & {\color{red}91.3} & {\color{red}83.2 }& {\color{red}72.2 }& {\color{red}55.6} & {\color{red}47.0 }\\
{\color{red}$\textrm{D}^2$$\textrm{L}^2$$\textrm{R}^2$~\cite{li2014learninglow}}  & {\color{red}94.5} & {\color{red}94.0} & {\color{red}89.5} & {\color{red} 86.5}& {\color{red}76.0} & {\color{red}70.5} \\
DLRD\_SR~\cite{ma2012sparse}      & 95.9          & 94.4          & 91.1          & 86.0          & 76.7          & 69.9          \\
FDDL~\cite{yang2011fisher}          & \textbf{96.7} & 94.0          & 89.8          & 85.1          & 76.1          & 68.3          \\
RPCA~\cite{candes2011robust}          & 89.3          & 88.0          & 83.0          & 76.6          & 72.0          & 66.2          \\
SRC~\cite{wright2009robust}           & 95.2          & 91.7          & 86.0          & 75.8          & 61.8          & 54.0          \\
SVM~\cite{ma2012sparse}           & 94.6          & 88.5          & 80.6          & 71.6          & 57.3          & 42.0          \\ \hline
\end{tabular}
\end{table}

\subsection{Unconstrained Face Classification}
\label{sec:5-5}
Thus far, we have conducted the experiments on  constrained datasets which do not exhibit large appearance variations of the same identity. In unconstrained scenarios face images of the same subject change dramatically due to variation in pose, illumination, expression and occlusion. Furthermore, test images may not contain the same type of variations and occlusions as the training images.
To evaluate the robustness of our method in unconstrained scenarios, we conduct experiments on the LFW database. This dataset contains images of 5,749 individuals and we use the LFW-a, which is an aligned version of LFW obtained using a commercial face alignment software. We select the subjects that include no less than ten samples and we construct a dataset with 158 subjects from LFW-a. For each person, we randomly select 5 samples for training (resulting in a dictionary of 790 faces) and the other 5 for testing. The images are resized to 90$\times$90.
\begin{table}[]
\centering
\caption{Recognition accuracy in (\%) on the LFW-a dataset}
\label{table:LFW}
\begin{tabular}{cc}
\hline
Methods & Accuracy \\ \hline
\textbf{RBDS}    & {\color{red}\textbf{71.7}}     \\
LRRS    &  62.1    \\
{\color{red}BDLRR~\cite{zhang2018discriminative}} & {\color{red}68.6} \\
{\color{red}$\textrm{D}^2$$\textrm{L}^2$$\textrm{R}^2$~\cite{li2014learninglow}}  & {\color{red}68.6} \\
SLRR~\cite{zhang2013learning}    &  68.2    \\
LRSI~\cite{wei2014robust}    &  66.2    \\
RPCA~\cite{candes2011robust}    &  66.3    \\
CRC~\cite{zhang2011sparse}     &  64.6    \\
SRC~\cite{wright2009robust}     &  68.3    \\
LLC~\cite{wang2010locality}     &  60.1    \\
SVM     &  58.2    \\ \hline
\end{tabular}
\end{table}
The experimental results are shown in Table~\ref{table:LFW}, where we can observe that our approach is superior to the competing methods. This again testifies to the effectiveness of RBDS.

\subsection{{\color{red}Scene} Categorization}
\label{sec:5-6}
The last experiment is performed on the Scene 15 dataset~\cite{lazebnik2006beyond}. This dataset contains 4485 images in total of 15 categories of natural scenes. Each class has 200 to 400 images, and the average image size is about 250$\times$300 pixels. This database consists a variety of outdoor and indoor scenes, such as office, kitchen, tall building and country scenes. The 3000-dimensional SIFT-based features provided in~\cite{jiang2013label} are exploited in our experiments. Following the common experimental setting used in~\cite{jiang2013label} and~\cite{lazebnik2006beyond}, 100 images per class are randomly selected as training data and the remaining images are used for testing.
\begin{table}[]
\centering
\caption{Recognition accuracy in (\%) achieved on the Scene 15 dataset}
\label{table:scene15}
\begin{tabular}{cc}
\hline
Methods & Accuracy \\ \hline
\textbf{RBDS}    & \textbf{98.66}     \\
LRRS    & 95.54    \\
{\color{red}BDLRR~\cite{zhang2018discriminative}} & {\color{red}98.50} \\
{\color{red}$\textrm{D}^2$$\textrm{L}^2$$\textrm{R}^2$~\cite{li2014learninglow}}  & {\color{red}96.58} \\
SLRR~\cite{zhang2013learning}    & 92.90    \\
LRSI~\cite{wei2014robust}    & 92.46    \\
RPCA~\cite{candes2011robust}    & 89.10     \\
CRC~\cite{zhang2011sparse}     & 92.00     \\
SRC~\cite{wright2009robust}     & 91.80     \\
LLC~\cite{wang2010locality}     & 89.20     \\
SVM     & 95.06    \\ \hline
\end{tabular}
\end{table}
The comparative results of all the approaches are presented in Table~\ref{table:scene15}. It can be seen that the proposed RBDS has the best performance. Note that compared with {\color{red}the recently proposed BDLRR method}, RBDS achieves a modest {\color{red}0.16}\% improvement.

\section{Conclusion}
\label{sec_6}
In this paper, we presented a low rank based method to learn image representations promoted by a block-diagonal structure constraint, \ie, RBDS, for pattern classification. A regularization term is incorporated into the framework of LRR to capture structure information globally. With this term, the off-block-diagonal elements of the representation matrix are minimized. As a result, the correlations between distinct classes are reduced while the coherence of intraclass representation is boosted. 
A compact dictionary is learned as part of the training process. We also proposed an effective algorithm to solve the optimization problem defined by our novel formulation. The experimental results obtained on five public datasets show that RBDS offers better recognition performance on average, and it is robust to appearance variations in illumination, expression, occlusion and random pixel corruption.

\begin{acknowledgement}

\end{acknowledgement}



\end{document}